\pdfoutput=1

\documentclass[11pt]{article}

\usepackage[]{EMNLP2023}

\usepackage{times}
\usepackage{latexsym}

\usepackage[T1]{fontenc}

\usepackage[utf8]{inputenc}

\usepackage{microtype}
\usepackage{multirow}
\usepackage{makecell}

\usepackage{inconsolata}
\usepackage{amsmath}
\usepackage{threeparttable}
\usepackage{booktabs}
\usepackage{url}
\usepackage{tablefootnote}
\usepackage{amsfonts}
\usepackage{graphicx}
\usepackage{xcolor}
\definecolor{mygrey}{gray}{0.5}

\newcommand{\figref}[1]{Figure \ref{#1}}
\newcommand{\eqnref}[1]{Eq. (\ref{#1})}
\newcommand{\tabref}[1]{Table \ref{#1}}

\title{DiffuSeq-v2: Bridging Discrete and Continuous Text Spaces\\ for Accelerated Seq2Seq Diffusion Models}

\author{Shansan Gong$^1$\quad Mukai Li$^1$\quad Jiangtao Feng$^2$\quad Zhiyong Wu$^3$\quad Lingpeng Kong$^1$ \\
$^1$The University of Hong Kong \quad $^2$Independent Researcher \quad $^3$Shanghai AI Lab \\
  \texttt{sansa933@connect.hku.hk jiangtaofeng0906@gmail.com lpk@cs.hku.hk}}

\begin{document}
\maketitle
\begin{abstract}
Diffusion models have gained prominence in generating high-quality sequences of text. Nevertheless, current approaches predominantly represent discrete text within a continuous diffusion space, which incurs substantial computational overhead during training and results in slower sampling speeds. In this paper, we introduce a soft absorbing state that facilitates the diffusion model in learning to reconstruct discrete mutations based on the underlying Gaussian space, thereby enhancing its capacity to recover conditional signals. During the sampling phase, we employ state-of-the-art ODE solvers within the continuous space to expedite the sampling process. Comprehensive experimental evaluations reveal that our proposed method effectively accelerates the training convergence by 4x and generates samples of similar quality 800x faster, rendering it significantly closer to practical application. \footnote{The code is released at \url{https://github.com/Shark-NLP/DiffuSeq/tree/diffuseq-v2}.}
\end{abstract}

\section{Introduction}
After diffusion models gained significant attention in the vision domain~\citep{ho2020denoising}, progress has been made in applying them to text generation tasks, including constrained text generation in Diffusion-LM~\citep{li2022diffusion} and sequence-to-sequence (Seq2Seq) text generation in DiffuSeq~\citep{gong2022diffuseq}. Numerous subsequent studies demonstrate that diffusion models have achieved results comparable to traditional autoregressive models and non-autoregressive models in tasks such as machine translation~\citep{Yuan2022SeqDiffuSeqTD, gao2022difformer, Zheng2023ARD} and summarization~\citep{Lin2022TextGW, Zhang2023DiffuSumGE, Mahabadi2023TESSTS, Zhou2023DiffusionNATSD}. However, most of these works suffer from slow convergence during training and slow generation speed, particularly considering that these approaches require the Minimum Bayes Risk (MBR) decoding strategy~\citep{koehn2004statistical} to enhance generation quality, resulting in a doubling of time consumption.

  

\begin{figure}
  \centering
  \includegraphics[width=0.46\textwidth]{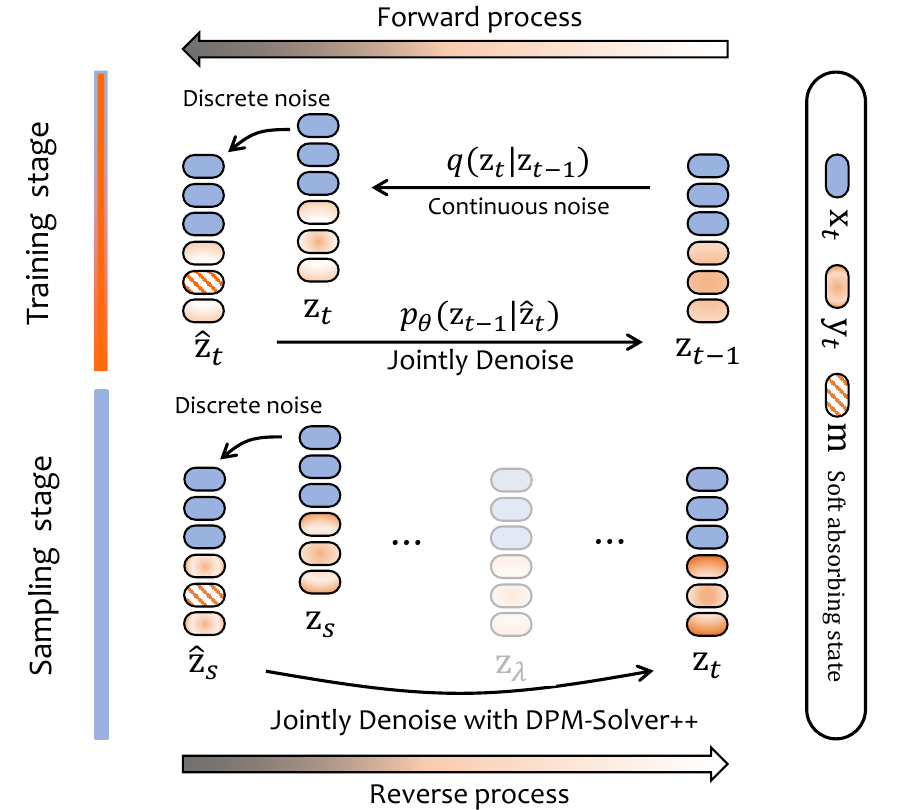}
  \caption{Training and sampling stages with discrete noise, which helps the two stages align better.}
  \label{fig:diffuseq++}
\end{figure}

In order to further narrow the gap between diffusion models and the prevailing autoregressive models, we aim to propose an accelerated version of DiffuSeq for both the training and sampling stages. For training, in addition to GPU acceleration techniques such as FP16, an improved training scheme can enable the model to better represent knowledge and learn data distribution more quickly~\citep{Hang2023EfficientDT}. For sampling, a well-trained diffusion model can achieve similar quality within a single sampling and without the need for MBR decoding, thus saving generation time. Furthermore, we can borrow the state-of-the-art ODE sampler DPM-solver++~\citep{lu2022dpm, lu2022dpm++} which is already applied in fast image generation. 
The progress in discrete text diffusion models~\citep{Zheng2023ARD} exhibits its superiority in using fewer sampling steps, thus intuitively inspiring us to bridge the gap between continuous and discrete spaces.

Based on the BERT~\citep{zhang2019bertscore} and BART~\citep{lewis2020bart}, as well as the absorbing state in D3PM~\citep{austin2021structured}, we propose incorporating an extra learned soft absorbing state and discretely adding it with Gaussian noise to jointly denoise the noise from two sources. The processes are illustrated in~\figref{fig:diffuseq++}.
Specifically, after posing Gaussian noise, we randomly replace the continuous vector of the sequence with the absorbing state. The ratio is set according to the time step. This approach bridges the gap between continuous and discrete diffusion processes and also makes the training and inference stages better aligned. It also facilitates the integration of the DPM-solver++ and reduces the number of required sampling steps.
In summary, our contributions are:

\begin{enumerate}

\item We introduce the learned soft absorbing state to help continuous diffusion models converge faster and eliminate the need for MBR decoding to ensure quality during sampling.

\item We adapt the DPM-solver++ to our enhanced diffusion text generation approach and demonstrate its feasibility in accelerating the generation speed experimentally.


\end{enumerate}

\section{Preliminaries}
\subsection{Continuous Diffusion Models}
\citet{ho2020denoising} and \citet{song2020denoising} formulate diffusion models in continuous space including forward and reverse processes. The forward process gradually corrupts data point $\mathbf{x}_0$ into a standard Gaussian noise $\mathbf{x}_T \sim \mathcal{N}(0, \mathbf{I})$. For each forward step $t \in [1, 2,...,T]$, the perturbation is followed by $q(\mathbf{x}_{t} \vert \mathbf{x}_{t-1}) = \mathcal{N}(\mathbf{x}_{t};\sqrt{1-\beta_t}\mathbf{x}_{t-1}, {\beta}_t \mathbf{I})$, with $\beta_t \in (0,1)$ as different scales. After the forward process, the reverse denoising process tries to gradually reconstruct the original data $\mathbf{x}_0$ via sampling from $\mathbf{x}_T$ by learning a diffusion model $f_{\theta}(\mathbf{x}_{t}, t)$. Diffusion-LM~\citep{li2022diffusion} and DiffuSeq~\citep{gong2022diffuseq} design an embedding function $\textsc{Emb}(\mathbf{w})$ to map the discrete text $\mathbf{w}$ into a continuous space and operate clamping on $\mathbf{x}_t$ to map it back to word embedding space at each sampling step to reduce rounding errors. 
\subsection{Discrete Diffusion Models}
For discrete diffusion probabilistic models, each $\mathbf{x}_t$ is a discrete random variable as one-hot vectors in $\{0, 1\}^K$, indicating the current state of each token, where $K$ is the size of the vocabulary. Multinomial diffusion~\citep{ho2020denoising} adopts a uniform noise distribution over the vocabulary. D3PM~\citep{austin2021structured} specifies $q(\mathbf{x}_{t} \vert \mathbf{x}_{t-1})$ through a transition matrix, and makes it to be a point mass with the probability on an absorbing state [MASK]. \citet{Zheng2023ARD} further derive an equivalent reparameterization to the discrete diffusion process. The resulting formulation is more amenable to training and leads to much-improved generation quality. 
However, discrete diffusion models may miss the opportunity to directly leverage the existing techniques from continuous diffusion models.

\begin{table*}[!th]
\setlength{\tabcolsep}{3.8pt}
\centering
\small
\caption{Sequence-to-sequence text generation results on QQP. All results are reported without MBR decoding if not specified. The best result is bolded, and the gray columns are excluded from comparison considering the fairness. The sampling step of BG-DiffuSeq is 20. The relative improvement $\uparrow$ is computed between our speedup version (step=2) and the DiffuSeq (MBR=1).}
\label{tb:main}
\begin{tabular}{lccccccccccc}
\toprule
     & \multicolumn{3}{c}{Continuous space}& \multicolumn{4}{c}{Discrete Space (step=10)}    & \multicolumn{4}{c}{Mixed space (ours)} \\ 
     \cmidrule(lr){2-4}
     \cmidrule(lr){5-8}
     \cmidrule(lr){9-12}
     &  \makecell[c]{\footnotesize{DiffuSeq}\\\tiny{(MBR=1)}} &  \textcolor{mygrey}{\makecell[c]{\footnotesize{DiffuSeq}\\\tiny{(MBR=10)}}} &  \makecell[c]{\footnotesize{BG}-\\\tiny{DiffuSeq}} &  \makecell[c]{\footnotesize{Multi}-\\ \footnotesize{nomial}} &  \textcolor{mygrey}{\makecell[c]{\footnotesize{RDM}+\\\tiny{Multinomial}}} &   \makecell[c]{\footnotesize{D3PM}-\\\tiny{absorbing}} & \textcolor{mygrey}{\makecell[c]{\footnotesize{RDM}+\\\tiny{absorbing}}} &  \textcolor{mygrey}{\makecell[c]{\footnotesize{Original}\\\tiny{(step=2000)}}} &  \makecell[c]{\footnotesize{Speedup}\\\tiny{(step=10)}} & \makecell[c]{\footnotesize{Speedup}\\\tiny{(step=2)}} & $\uparrow$\\ 
     \midrule
BLEU      & 0.1884 & \textcolor{mygrey}{0.2413} & 0.2077    &  0.2025 &  \textcolor{mygrey}{0.2315}   & \textbf{0.2245} &  \textcolor{mygrey}{0.2313}   &  \textcolor{mygrey}{0.2411}  & 0.2210 & 0.2115 & 12.3\%\\ 
R-L       & 0.5327 & \textcolor{mygrey}{0.5880}   & 0.5652   &  0.5516   &  \textcolor{mygrey}{0.5712}   & 0.5677 &  \textcolor{mygrey}{0.5735}   & \textcolor{mygrey}{0.5930}  & \textbf{0.5732} & 0.5651 & 6.1\%\\ 
BertScore & 0.7965 & \textcolor{mygrey}{0.8365}  & 0.8057    &  0.7966  & \textcolor{mygrey}{0.8375}  & \textbf{0.8281} &  \textcolor{mygrey}{0.8416}  & \textcolor{mygrey}{0.8393}  & 0.8207 & 0.8036 & 0.8\%\\ 
\midrule
Speed (it/s) &  0.51 & \textcolor{mygrey}{0.05}    &  60.3  &  106.4  & \textcolor{mygrey}{108.6}  & 116.1 &  \textcolor{mygrey}{121.0}   &   \textcolor{mygrey}{1.07}  & 208.3 & \textbf{406.5} & $\approx 800 \times$ \\

\bottomrule
\end{tabular}
\end{table*}

\section{Methods}
In DiffuSeq, it formulates Seq2Seq tasks as conditional generation and learns $p(\mathbf{w}^x|\mathbf{w}^y)$, where $\mathbf{w}^x$ and $\mathbf{w}^y$ refer to the input and target sequence separately. We follow the notation, concatenating two sequences as $\mathbf{z}_{t} = \mathbf{x}_t \oplus \mathbf{y}_t$, where $\mathbf{x}_t$ and $\mathbf{y}_t$ represent parts of $\mathbf{z}_{t}$ that belong to $\mathbf{w}^x$ and $\mathbf{w}^y$, respectively. To accelerate continuous diffusion models as well as leverage the absorbing state of discrete diffusion models, we add learnable soft absorbing state into DiffuSeq. We first combine the continuous Gaussian noise and discrete absorbing noise, and then jointly denoise them. The detailed derivations can be found in Appendix~\ref{sec:aderivations}.
\subsection{Training Stage}
\paragraph{Forward process with soft absorbing state} In continuous space, we first add Gaussian noise $\epsilon$ for each time step with $\alpha_t = 1-\beta_t$, $\bar{\alpha}_t = \prod_{i=1}^t \alpha_i$:
\begin{equation}
    \mathbf{z}_t = \sqrt{\bar{\alpha_t}}\mathbf{z}_0+\sqrt{1-\bar{\alpha}_t}\epsilon.
\end{equation}
Considering the $i$-th token in the hidden representation $\mathbf{z}_t$ of the word sequence, we replace its representation with our soft absorbing state $\mathbf{m}$ at a certain probability. The soft absorbing state $\mathbf{m}$ is in the same hidden dimension as word embeddings and is also jointly learned along with the whole diffusion process.
\begin{equation}
\label{eq:z_it}
    \mathbf{\hat{z}}_t^i = 
    \begin{cases}
    \mathbf{m}&\mbox{if $\rho=1$}\\
    \mathbf{z}_t^i&\mbox{if $\rho=0$}
    \end{cases},
\end{equation}
where $\rho=\mathtt{Bernoulli}(\beta_t * \gamma)$, and $\gamma$ is the [MASK] ratio when $t=T$. This operation keeps the diffusion model in continuous space but discretely replaces the representation of some tokens in the sequence, and the replacement probability is also scaled to the time step, same with $\beta_t$. Noted that these two kinds of noise are posed partially on the target $\mathbf{y}_t$ space in the manner of DiffuSeq.
\paragraph{Jointly denoise} The reverse process is to jointly reconstruct the corrupted data point. The simplified loss function is almost the same with DiffuSeq, except for $\mathbf{z}_t$ in different noise strategies:
\begin{equation}
\small
\label{eq:loss}
\begin{aligned}
    \min_{\theta}\; \mathcal{L}_{\text{VLB}} & = \min_{\theta} \left[ \sum_{t=2}^T||\mathbf{y}_0-\tilde f_{\theta}(\mathbf{\hat{z}}_t, t)||^2 + \right.\\
    & \left. {||\textsc{Emb}(\mathbf{w}^y)-\tilde f_{\theta}(\mathbf{\hat{z}}_1, 1)||^2} + \mathcal{R}(||\mathbf{z}_0||^2) \right].
\end{aligned}
\end{equation}

\subsection{Sampling Stage}
Previous continuous text diffusion models adopt clamp operation to make the vector predictions more precise and reduce rounding errors~\citep{li2022diffusion} during sampling. However, this operation is not deployed in the training stage, and the gap between training and sampling~\citep{Tang2023CanDM} may hinder the performance and the further optimization for sampling speed. 

On the contrary, for our methods, during sampling, the same discrete noise in \eqnref{eq:z_it} is sprinkled in the continuous Gaussian noise, which bridges the training and inference in discrete space. Using the exact solution of diffusion ODEs proposed by DPM-solver++~\citep{lu2022dpm, lu2022dpm++}, given an initial value $\mathbf{z}_s$ at time $s > 0$, we have:
\begin{equation}
    \mathbf{z}_t = \frac{\sigma_t}{\sigma_s} \mathbf{z}_s + \sigma_t \int_{\lambda_s}^{\lambda_t} e^{\lambda} f_\theta(\mathbf{\hat{z}}_\lambda, \lambda) \,d\lambda,
\end{equation}
where the $\lambda$ is a strictly decreasing function of $t$, $\sigma_t$ is monotonic to $\beta_t$, and $f_\theta(\mathbf{\hat{z}}_\lambda, \lambda)$ is aligned with training objectives. The integral term can be analytically computed by repeatedly applying integration-by-parts $n$ times, and we can just approximate the first several orders and drop the high-order error terms. We use second order in the experiment.

\begin{figure*}[!tb]
\centering
\begin{minipage}[t]{0.3\textwidth}
\centering
\includegraphics[height=3cm]{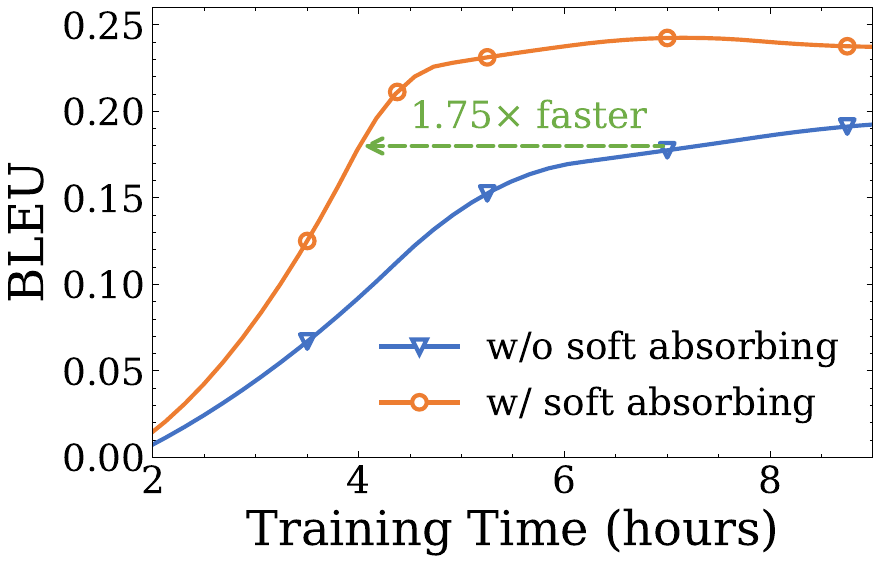}
\caption{The test BLEU score along with training hours under different training schemes.}
\label{fig:training}
\end{minipage}
\hspace{0.2cm}
\begin{minipage}[t]{0.3\textwidth}
\centering
\includegraphics[height=3cm]{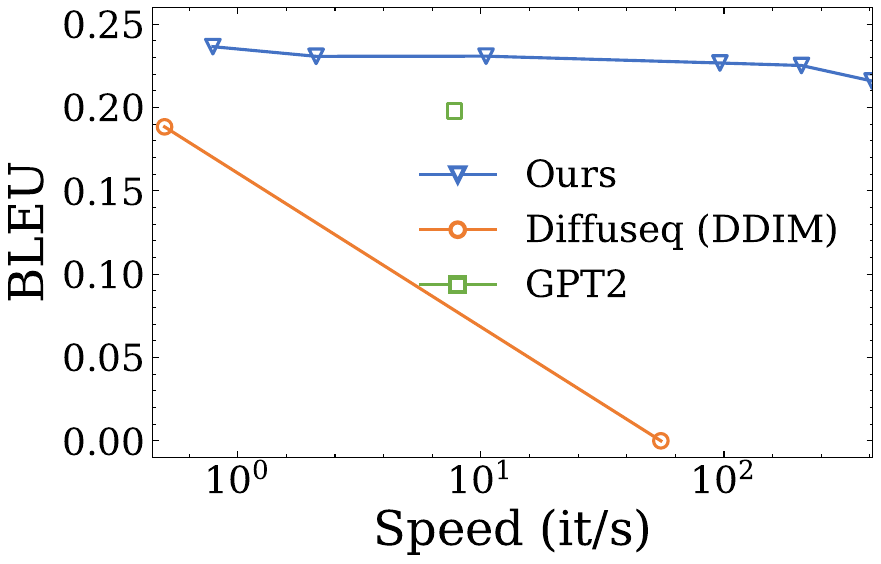}
\caption{Generation speed and quality under different sampling steps incorporating DPM-solver.}
\label{fig:dpm-solver}
\end{minipage}
\hspace{0.2cm}
\begin{minipage}[t]{0.3\textwidth}
\centering
\includegraphics[height=3cm]{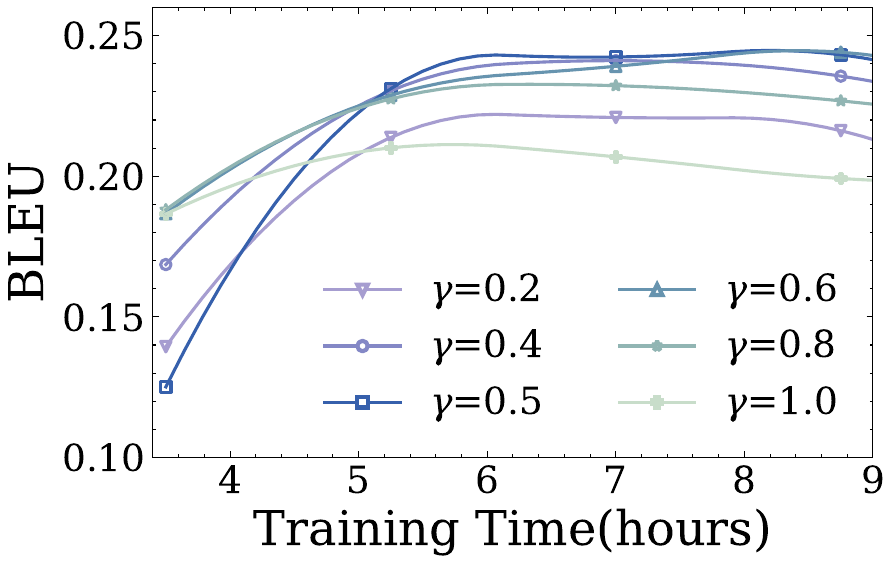}
\caption{The test BLEU score at different training hours for different settings of the ratio $\gamma$.}
\label{fig:denoise_rate}
\end{minipage}
\end{figure*}

\section{Experiments}
We try to validate two main research questions. RQ1: Can the soft absorbing state added in continuous diffusion models improve the generation quality and boost the training convergence? RQ2: To what extent does the DPM ODE solver improve sampling speed and affect the performance?

\subsection{Experiment setup}
\paragraph{Dataset}
We adopt QQP
\footnote{\url{https://www.kaggle.com/c/quora-question-pairs}}
 for paraphrasing the sentence to the same semantic content, which is lightweight to train and has been widely used in many Seq2Seq text diffusion models.
\paragraph{Baselines}
We choose DiffuSeq~\citep{gong2022diffuseq}, BG-DiffuSeq~\citep{Tang2023CanDM} as the representative of continuous diffusion models. The latter is the enhanced version of DiffuSeq, targeting bridging the gap between training and sampling to generate high-quality texts within fewer steps. We choose multinomial~\citep{hoogeboom2021argmax}, D3PM-absorbing~\citep{austin2021structured}, and a reparameterized version of them~\citep{Zheng2023ARD} as the representative of discrete diffusion models, with fewer diffusion steps for training.
\paragraph{Implementation details}
We follow the training, sampling, and evaluation implementation of DiffuSeq\footnote{\url{https://github.com/Shark-NLP/DiffuSeq}}. The experiment is deployed on NVIDIA A100 80G GPUs, with 2 GPUs for training. More details can be seen in Appendix~\ref{sec:appendix details}.
\subsection{Main results}
As seen from \tabref{tb:main}, original DiffuSeq relies on the MBR decoding to ensure generation quality. By contrast, our method in the original sampling version achieves comparable performance without applying MBR by sampling 10 times, which significantly saves the time for generating high-quality texts. Even the speedup version (step=2) still outperforms DiffuSeq (MBR=1) a lot, according to the relative improvement. Our method is also superior to BG-DiffuSeq, which bridges the gap between training and sampling in continuous space. The speedup versions of our methods still have an advantage over discrete diffusion models, especially in terms of generation speed. We do not directly compare with RDMs because it designs an algorithm to route the discrete change of words, while ours just uses vanilla transition which uniformly changes tokens to the soft absorbing state. 


\subsection{Training speed}
We use FP16 for GPU acceleration~\citep{ott-etal-2018-scaling}, which reduces the total training time from 28 to 11 hours with 2 GPUs and brings $2.5\times$ speed up. 
According to~\figref{fig:training}, the jointly denoise training scheme expedites training convergence by at least $1.75\times$, probably because the absorbing state perturbs the representation of sequence discretely, which empowers the model with better capacity to reconstruct the discrete text information. The training consumption is saved more than $4\times$ in total. 
\subsection{Sampling speed}
Sampling with FP16 is approximately $2\times$ faster than the original DiffuSeq. Furthermore, the incorporation of DPM-solver++ shrinks the sampling step to 10 or even 2 without sacrificing the performance much, as shown in~\figref{fig:dpm-solver}. This improvement is significant compared with DDIM~\cite{nichol2021improved} used in DiffuSeq. Comparing our speedup version (step=2) with DiffuSeq (MBR=1), texts with higher quality are generated conditionally and meanwhile about 800$\times$ faster.
\subsection{Ablation study}
We test different sampling strategies in~\tabref{tb:sampling}. After removing the clamp operation, the performance of DiffuSeq gets affected evidently, while ours seldom drop, which validates our previous assumption that adding the soft absorbing state bridges the gap between training and sampling, and we can remove the clamp operation. 
And further plugging in with DPM-solver++ will not introduce extra rounding errors. 
We test sampling without soft absorbing state on our model which still adds this discrete noise in the training stage.
A significant drop can be seen, demonstrating the importance of the alignment of training and sampling.
We also analyze the choice of the hyper-parameter $\gamma$. As seen in~\figref{fig:denoise_rate}, too small or too large [MASK] rate will harm the performance, and closer to the middle tends to perform better. So we choose $\gamma=0.5$ as the default setting in our experiment. 

\begin{table}[t]
\setlength{\tabcolsep}{4pt}
\centering
\small
\caption{Different sampling strategies. [C] denotes clamp operation and w/o [M] denotes stopping adding discrete noise during sampling.}
\label{tb:sampling}
\begin{tabular}{lccccc}
\toprule
     & \multicolumn{2}{c}{DiffuSeq}& \multicolumn{3}{c}{Ours (Original)} \\ 
     \cmidrule(lr){2-3}
     \cmidrule(lr){4-6}
     &  w [C] & w/o [C] & w [C] & w/o [C] & w/o [M]\\ 
     \midrule
BLEU      & 19.87 & 19.19 & 24.11 & 23.98 & 20.50 \\
Drop (\%)      & -& 3.4 & - & 0.5 & 14.9 \\

\bottomrule
\end{tabular}

\end{table}
\section{Conclusions}
In this work, we present a simple but effective training scheme for joint discrete and continuous text diffusion models, which additionally resets some tokens into the soft absorbing state. The discrete noise bridges the training and sampling stages, saving time consumption of these two stages and the plugged DPM-solver++ further makes the sampling faster. Our method is orthogonal to many other techniques such as self-conditioning~\cite{Mahabadi2023TESSTS, chen2022analog}, choosing tokens with different importance~\cite{he2022diffusionbert}, which can also further enhance generation quality. Our method is fundamental to diffusion text generation and can be applied beyond DiffuSeq~\citep{chen2023cheaper}.

\section*{Limitations}
Regarding the methods, we opt not to incorporate length prediction, unlike other approaches. Instead, we use the [PAD] token to indicate the length automatically. This may require more GPU memory for text generation.
We validate our methods on the QQP dataset, which is one of the sequence-to-sequence text generation tasks. However, due to resource and time constraints, we are unable to test the effectiveness of our methods on more complex tasks such as machine translation and summarization. Additionally, this work does not explore the impact of scaling up the model size.

\section*{Ethics Statement}
In this study, we used diffusion models to generate text. To ensure that the generated text is free from bias, we carefully selected our training data and evaluated the quality of the generated text. We acknowledge that our research has potential ethical implications, particularly in the area of text generation for malicious purposes. To mitigate this risk, we have taken steps to ensure that our research is conducted in an ethical manner and that the generated text is used only for legitimate purposes.


\bibliography{anthology,custom}

\begin{thebibliography}{29}
\expandafter\ifx\csname natexlab\endcsname\relax\def\natexlab#1{#1}\fi

\bibitem[{Austin et~al.(2021)Austin, Johnson, Ho, Tarlow, and van~den
  Berg}]{austin2021structured}
Jacob Austin, Daniel~D Johnson, Jonathan Ho, Daniel Tarlow, and Rianne van~den
  Berg. 2021.
\newblock Structured denoising diffusion models in discrete state-spaces.
\newblock \emph{Advances in Neural Information Processing Systems}.

\bibitem[{Chen et~al.(2023)Chen, Zhang, Li, Smola, and Yang}]{chen2023cheaper}
Jiaao Chen, Aston Zhang, Mu~Li, Alex Smola, and Diyi Yang. 2023.
\newblock A cheaper and better diffusion language model with soft-masked noise.
\newblock \emph{ArXiv}, abs/2304.04746.

\bibitem[{Chen et~al.(2022)Chen, Zhang, and Hinton}]{chen2022analog}
Ting Chen, Ruixiang Zhang, and Geoffrey Hinton. 2022.
\newblock Analog bits: Generating discrete data using diffusion models with
  self-conditioning.
\newblock \emph{ArXiv}, abs/2208.04202.

\bibitem[{Gao et~al.(2022)Gao, Guo, Tan, Zhu, Zhang, Bian, and
  Xu}]{gao2022difformer}
Zhujin Gao, Junliang Guo, Xu~Tan, Yongxin Zhu, Fang Zhang, Jiang Bian, and
  Linli Xu. 2022.
\newblock Difformer: Empowering diffusion model on embedding space for text
  generation.
\newblock \emph{ArXiv}, abs/2212.09412.

\bibitem[{Gong et~al.(2023)Gong, Li, Feng, Wu, and Kong}]{gong2022diffuseq}
Shansan Gong, Mukai Li, Jiangtao Feng, Zhiyong Wu, and Lingpeng Kong. 2023.
\newblock {DiffuSeq}: Sequence to sequence text generation with diffusion
  models.
\newblock In \emph{International Conference on Learning Representations, ICLR}.

\bibitem[{Hang et~al.(2023)Hang, Gu, Li, Bao, Chen, Hu, Geng, and
  Guo}]{Hang2023EfficientDT}
Tiankai Hang, Shuyang Gu, Chen Li, Jianmin Bao, Dong Chen, Han Hu, Xin Geng,
  and Baining Guo. 2023.
\newblock Efficient diffusion training via min-snr weighting strategy.
\newblock \emph{ArXiv}, abs/2303.09556.

\bibitem[{He et~al.(2022)He, Sun, Wang, Huang, and Qiu}]{he2022diffusionbert}
Zhengfu He, Tianxiang Sun, Kuanning Wang, Xuanjing Huang, and Xipeng Qiu. 2022.
\newblock Diffusionbert: Improving generative masked language models with
  diffusion models.
\newblock \emph{ArXiv}, abs/2211.15029.

\bibitem[{Ho et~al.(2020)Ho, Jain, and Abbeel}]{ho2020denoising}
Jonathan Ho, Ajay Jain, and Pieter Abbeel. 2020.
\newblock Denoising diffusion probabilistic models.
\newblock \emph{Advances in Neural Information Processing Systems}.

\bibitem[{Hoogeboom et~al.(2021)Hoogeboom, Nielsen, Jaini, Forr{\'e}, and
  Welling}]{hoogeboom2021argmax}
Emiel Hoogeboom, Didrik Nielsen, Priyank Jaini, Patrick Forr{\'e}, and Max
  Welling. 2021.
\newblock Argmax flows and multinomial diffusion: Learning categorical
  distributions.
\newblock \emph{Advances in Neural Information Processing Systems}.

\bibitem[{Hu et~al.(2022)Hu, Zheng, Zheng, Cham, Wang, Yang, Tao, and
  Suganthan}]{hu2022unified}
Minghui Hu, Chuanxia Zheng, Heliang Zheng, Tat-Jen Cham, Chaoyue Wang, Zuopeng
  Yang, Dacheng Tao, and Ponnuthurai~N Suganthan. 2022.
\newblock Unified discrete diffusion for simultaneous vision-language
  generation.
\newblock \emph{ArXiv}, abs/2211.14842.

\bibitem[{Koehn(2004)}]{koehn2004statistical}
Philipp Koehn. 2004.
\newblock Statistical significance tests for machine translation evaluation.
\newblock In \emph{Proceedings of the 2004 conference on empirical methods in
  natural language processing}, pages 388--395.

\bibitem[{Lewis et~al.(2020)Lewis, Liu, Goyal, Ghazvininejad, Mohamed, Levy,
  Stoyanov, and Zettlemoyer}]{lewis2020bart}
Mike Lewis, Yinhan Liu, Naman Goyal, Marjan Ghazvininejad, Abdelrahman Mohamed,
  Omer Levy, Veselin Stoyanov, and Luke Zettlemoyer. 2020.
\newblock Bart: Denoising sequence-to-sequence pre-training for natural
  language generation, translation, and comprehension.
\newblock In \emph{Proceedings of the 58th Annual Meeting of the Association
  for Computational Linguistics}, pages 7871--7880.

\bibitem[{Li et~al.(2022)Li, Thickstun, Gulrajani, Liang, and
  Hashimoto}]{li2022diffusion}
Xiang~Lisa Li, John Thickstun, Ishaan Gulrajani, Percy Liang, and Tatsunori~B
  Hashimoto. 2022.
\newblock Diffusion-lm improves controllable text generation.
\newblock \emph{ArXiv}, abs/2205.14217.

\bibitem[{Lin(2004)}]{lin2004rouge}
Chin-Yew Lin. 2004.
\newblock Rouge: A package for automatic evaluation of summaries.
\newblock In \emph{Text summarization branches out}.

\bibitem[{Lin et~al.(2022)Lin, Gong, Shen, Wu, Fan, Lin, Duan, and
  Chen}]{Lin2022TextGW}
Zheng-Wen Lin, Yeyun Gong, Yelong Shen, Tong Wu, Zhihao Fan, Chen Lin, Nan
  Duan, and Weizhu Chen. 2022.
\newblock Text generation with diffusion language models: A pre-training
  approach with continuous paragraph denoise.
\newblock volume abs/2212.11685.

\bibitem[{Lu et~al.(2022{\natexlab{a}})Lu, Zhou, Bao, Chen, Li, and
  Zhu}]{lu2022dpm}
Cheng Lu, Yuhao Zhou, Fan Bao, Jianfei Chen, Chongxuan Li, and Jun Zhu.
  2022{\natexlab{a}}.
\newblock Dpm-solver: A fast ode solver for diffusion probabilistic model
  sampling in around 10 steps.
\newblock In \emph{Conference on Neural Information Processing Systems,
  NeurIPS}.

\bibitem[{Lu et~al.(2022{\natexlab{b}})Lu, Zhou, Bao, Chen, Li, and
  Zhu}]{lu2022dpm++}
Cheng Lu, Yuhao Zhou, Fan Bao, Jianfei Chen, Chongxuan Li, and Jun Zhu.
  2022{\natexlab{b}}.
\newblock Dpm-solver++: Fast solver for guided sampling of diffusion
  probabilistic models.
\newblock \emph{ArXiv}, abs/2211.01095.

\bibitem[{Mahabadi et~al.(2023)Mahabadi, Tae, Ivison, Henderson, Beltagy,
  Peters, and Cohan}]{Mahabadi2023TESSTS}
Rabeeh~Karimi Mahabadi, Jaesung Tae, Hamish Ivison, James Henderson,
  Iz~Beltagy, Matthew~E. Peters, and Arman Cohan. 2023.
\newblock Tess: Text-to-text self-conditioned simplex diffusion.
\newblock \emph{ArXiv}, abs/2305.08379.

\bibitem[{Nichol and Dhariwal(2021)}]{nichol2021improved}
Alexander~Quinn Nichol and Prafulla Dhariwal. 2021.
\newblock Improved denoising diffusion probabilistic models.
\newblock In \emph{International Conference on Machine Learning, ICML}.

\bibitem[{Ott et~al.(2018)Ott, Edunov, Grangier, and
  Auli}]{ott-etal-2018-scaling}
Myle Ott, Sergey Edunov, David Grangier, and Michael Auli. 2018.
\newblock Scaling neural machine translation.
\newblock In \emph{Proceedings of the Third Conference on Machine Translation:
  Research Papers}, pages 1--9. Association for Computational Linguistics.

\bibitem[{Papineni et~al.(2002)Papineni, Roukos, Ward, and
  Zhu}]{papineni-etal-2002-bleu}
Kishore Papineni, Salim Roukos, Todd Ward, and Wei-Jing Zhu. 2002.
\newblock {B}leu: a method for automatic evaluation of machine translation.
\newblock In \emph{Proceedings of the 40th Annual Meeting of the Association
  for Computational Linguistics, ACL}.

\bibitem[{Song et~al.(2020)Song, Meng, and Ermon}]{song2020denoising}
Jiaming Song, Chenlin Meng, and Stefano Ermon. 2020.
\newblock Denoising diffusion implicit models.
\newblock In \emph{International Conference on Learning Representations, ICLR}.

\bibitem[{Tang et~al.(2023)Tang, Wang, Zhou, Li, Cao, and
  Zhang}]{Tang2023CanDM}
Zecheng Tang, Pinzheng Wang, Keyan Zhou, Juntao Li, Ziqiang Cao, and M.~Zhang.
  2023.
\newblock Can diffusion model achieve better performance in text generation?
  bridging the gap between training and inference!
\newblock \emph{ArXiv}, abs/2305.04465.

\bibitem[{Yuan et~al.(2022)Yuan, Yuan, Tan, Huang, and
  Huang}]{Yuan2022SeqDiffuSeqTD}
Hongyi Yuan, Zheng Yuan, Chuanqi Tan, Fei Huang, and Songfang Huang. 2022.
\newblock Seqdiffuseq: Text diffusion with encoder-decoder transformers.
\newblock \emph{ArXiv}, abs/2212.10325.

\bibitem[{Zhang et~al.(2023)Zhang, Liu, and Zhang}]{Zhang2023DiffuSumGE}
Haopeng Zhang, Xiao Liu, and Jiawei Zhang. 2023.
\newblock Diffusum: Generation enhanced extractive summarization with
  diffusion.
\newblock \emph{ArXiv}, abs/2305.01735.

\bibitem[{Zhang et~al.(2019)Zhang, Kishore, Wu, Weinberger, and
  Artzi}]{zhang2019bertscore}
Tianyi Zhang, Varsha Kishore, Felix Wu, Kilian~Q Weinberger, and Yoav Artzi.
  2019.
\newblock Bertscore: Evaluating text generation with bert.
\newblock In \emph{International Conference on Learning Representations, ICLR}.

\bibitem[{Zheng et~al.(2023)Zheng, Yuan, Yu, and Kong}]{Zheng2023ARD}
Lin Zheng, Jianbo Yuan, Lei Yu, and Lingpeng Kong. 2023.
\newblock A reparameterized discrete diffusion model for text generation.
\newblock \emph{ArXiv}, abs/2302.05737.

\bibitem[{Zhou et~al.(2023)Zhou, Li, Zhao, and rong
  Wen}]{Zhou2023DiffusionNATSD}
Kun Zhou, Yifan Li, Wayne~Xin Zhao, and Ji~rong Wen. 2023.
\newblock Diffusion-nat: Self-prompting discrete diffusion for
  non-autoregressive text generation.
\newblock \emph{ArXiv}, abs/2305.04044.

\bibitem[{Zhu et~al.(2018)Zhu, Lu, Zheng, Guo, Zhang, Wang, and
  Yu}]{zhu2018texygen}
Yaoming Zhu, Sidi Lu, Lei Zheng, Jiaxian Guo, Weinan Zhang, Jun Wang, and Yong
  Yu. 2018.
\newblock Texygen: A benchmarking platform for text generation models.
\newblock In \emph{The 41st International ACM SIGIR Conference on Research \&
  Development in Information Retrieval}, pages 1097--1100.

\end{thebibliography}
\bibliographystyle{acl_natbib}

\appendix
\section{Detailed Derivations of Our Methods}
\label{sec:aderivations}
\subsection{Training stage}
Following~\citet{ho2020denoising,nichol2021improved,song2020denoising,li2022diffusion, gong2022diffuseq}, we define the forward noising process and reverse denoising process on the latent continuous space $\mathbf{z}$. 

The \textit{forward} noising is to perturb the structure of data $\mathbf{z}_0$. $\mathbf{z}_0$ is finally changed into the partial Gaussian noise with $\mathbf{y}_T\sim \mathcal{N}(0, \mathbf{I})$ through $T$-step forward random disturbance 
\begin{equation}
    q(\mathbf{z}_{t} \vert \mathbf{z}_{t-1}) = \mathcal{N}(\mathbf{z}_{t};\sqrt{1-\beta_t}\mathbf{z}_{t-1}, {\beta}_t \mathbf{I}),
\end{equation}
with $t = 1, 2,...,T$ and $\{\beta_t \in (0,1)\}_{t=1}^T$ are the variance schedule. Let $\alpha_t=1-\beta_t$ and $\bar{\alpha}_t = \prod_{i=1}^t \alpha_i$, we have:
\begin{equation}
\begin{aligned}
\label{eq:zt}
    \mathbf{z}_t = &\sqrt{\alpha_t} \mathbf{z}_{t-1}+\sqrt{1-\alpha_t}\epsilon_{t-1}\\
    =&\sqrt{\alpha_t\alpha_{t-1}} \mathbf{z}_{t-2}+\sqrt{1-\alpha_t\alpha_{t-1}}\bar{\epsilon}_{t-2}\\
    =&...=\sqrt{\bar{\alpha_t}}\mathbf{z}_0+\sqrt{1-\bar{\alpha}_t}\epsilon,
\end{aligned}
\end{equation}
where $\epsilon$ stands for Gaussian noises. In the end, $q(\mathbf{z}_t \vert \mathbf{z}_0) = \mathcal{N}(\mathbf{z}_t; \sqrt{\bar{\alpha}_t} \mathbf{z}_0, (1 - \bar{\alpha}_t)\mathbf{I})$. We use a sqrt noise schedule $\bar{\alpha}_t=1-\sqrt{t/T+s}$ with $s$ as a small constant at the start of the noise level.

Based on $q(\mathbf{z}_{t} \vert \mathbf{z}_{t-1})$, we further add discrete noise. 
Consider the $i$-th token in the hidden representation $\mathbf{z}_t$ of the word sequence, we replace its representation with our soft absorbing state $\mathbf{m}$ at a certain probability. The soft absorbing state $\mathbf{m}$ is in the same hidden dimension as word embeddings and is also jointly learned along with the whole diffusion process.
\begin{equation}
    \mathbf{\hat{z}}_t^i = 
    \begin{cases}
    \mathbf{m}&\mbox{if $\rho=1$}\\
    \mathbf{z}_t^i&\mbox{if $\rho=0$}
    \end{cases},
\end{equation}
where $\rho=\mathtt{Bernoulli}(\beta_t * \gamma)$, and $\gamma$ is the [MASK] ratio when $t=T$. The replacement probability $\rho$ is also scaled to the time step, proportional to $\beta_t$. All these two kinds of noise are posed partially in the manner of DiffuSeq, to make sure the condition signal $\mathbf{x}$ unchanged.

For \textit{reverse} process, the continuous noise is accumulated as before, and defined as:
\begin{equation}
    p_{\theta}(\mathbf{z}_{0:T}):=p(\mathbf{z}_T)\prod_{t=1}^Tp_{\theta}(\mathbf{z}_{t-1}|\mathbf{z}_t),
\end{equation}
\begin{equation}
    p_{\theta}(\mathbf{z}_{t-1}|\mathbf{z}_t)=\mathcal{N}(\mathbf{z}_{t-1};\mu_{\theta}(\mathbf{z}_t, t), \sigma_{\theta}(\mathbf{z}_t, t)),
\end{equation}
where the $\mu_{\theta}(\cdot)$ and $\sigma_{\theta}(\cdot)$ is the predicted parameterization of the mean and standard variation of $ q(\mathbf{z}_t|\mathbf{z}_{t-1})$ in forward process. Using Bayes' rule:
\begin{equation}
\small
\begin{aligned}
    q(\mathbf{z}_{t-1} \vert \mathbf{z}_t, \mathbf{z}_0) 
    &= q(\mathbf{z}_t \vert \mathbf{z}_{t-1}, \mathbf{z}_0) \frac{ q(\mathbf{z}_{t-1} \vert \mathbf{z}_0) }{ q(\mathbf{z}_t \vert \mathbf{z}_0) },
\end{aligned}
\end{equation}
substitute Eq.~(\ref{eq:zt}) to it and we can get the parameterized mean of $q(\mathbf{z}_{t-1} \vert \mathbf{z}_t, \mathbf{z}_0)$: 
\begin{equation}
\small
\label{eq:ut}
    \mu_t(\mathbf{z}_t,\mathbf{z}_0)=\frac{\sqrt{\alpha_t}(1-\bar{\alpha}_{t-1})}{1-\bar{\alpha}_t}\mathbf{z}_t+\frac{\sqrt{\bar{\alpha}_{t-1}}\beta_t}{1-\bar{\alpha}_t}\mathbf{z}_0,
\end{equation}
and for brevity, we moit the coefficient of $\mathbf{z}_t$ and $\mathbf{z}_0$ as constants.

We can use the variational lower bound to optimize the negative log-likelihood $\mathbb{E}[-\log p_{\theta}(\mathbf{x}_0) ]\leq \mathcal{L}_\text{VLB}$. The objective can be further rewritten to be a combination of several KL-divergence.
\begin{equation}
\small
\begin{aligned}
\mathcal{L}_\text{VLB} & =
\mathcal{L}_T + \mathcal{L}_{T-1} + \dots + \mathcal{L}_0 \\
&=\mathbb{E}_{ q(\mathbf{z}_{1:T}|\mathbf{z}_0)}
\Bigg[
\log\frac{ q(\mathbf{z}_T|\mathbf{z}_0)}{p_{\theta}(\mathbf{z}_T)}  \\
& + \sum_{t=2}^T \log{\frac{ q(\mathbf{z}_{t-1}|\mathbf{z}_0,\mathbf{z}_t)}{p_{\theta}(\mathbf{z}_{t-1}|\mathbf{z}_t)}} \\
& +  {\log\frac{q_{\phi}(\mathbf{z}_0|\mathbf{w}^{x\oplus y})}{p_{\theta}(\mathbf{z}_0|\mathbf{z}_1)}}-\log p_{\theta}(\mathbf{w}^{x\oplus y}|\mathbf{z}_0)\vphantom{\log{\frac{ q(\mathbf{z}_{t-1}|\mathbf{z}_0,\mathbf{z}_t)}{p_{\theta}(\mathbf{z}_{t-1}|\mathbf{z}_t)}}}
\Bigg].\\
\end{aligned}
\end{equation}
For $1 \leq t \leq T-1$, we compute the parameterization of $\mathcal{L}_t$ by substituting Eq.~(\ref{eq:ut}) to minimize the difference from $\mu_t$ and $\mu_{\theta}$ following~\citet{ho2020denoising}:
\begin{equation}
\begin{aligned}
\mathcal{L}_t&=\mathbb{E}_{\mathbf{z}_0}\left[\log{\frac{ q(\mathbf{z}_{t}|\mathbf{z}_0,\mathbf{z}_{t+1})}{p_{\theta}(\mathbf{z}_{t}|\mathbf{z}_{t+1})}}\right]\\ & ={\mathcal{C}}\mathbb{E}_{\mathbf{z}_0}[||\mathbf{z}_0-f_{\theta}(\mathbf{z}_t, t)||^2],  \\
\end{aligned}
\end{equation}
where $\mathcal{C}$ is a loss independent constant. 

It is intuitively believed that the above continuous diffusion models learns aims to recover the corrupt data $\mathbf{z}_t$ with $f_{\theta}$ to $\mathbf{z}_0$. When we look back to our added discrete noise, since its directly added to the $ q(\mathbf{z}_t | \mathbf{z}_0)$, which can be directly modeled by $f_{\theta}$. The $\mathcal{L}_t$ should be:
\begin{equation}
    \mathcal{L}_t = {\mathcal{C}}\mathbb{E}_{\mathbf{z}_0}[||\mathbf{z}_0-f_{\theta}(\mathbf{\hat{z}}_t, t)||^2].
\end{equation}

Then the optimization of training loss $\min_{\theta}\;\mathcal{L}_{\text{VLB}}$ can be further simplified as:
\begin{equation}
\small
\begin{aligned}
    \min_{\theta}\; \mathcal{L}_{\text{VLB}} & = \min_{\theta} \left[ \sum_{t=2}^T||\mathbf{y}_0-\tilde f_{\theta}(\mathbf{\hat{z}}_t, t)||^2 + \right.\\
    & \left. {||\textsc{Emb}(\mathbf{w}^y)-\tilde f_{\theta}(\mathbf{\hat{z}}_1, 1)||^2} + \mathcal{R}(||\mathbf{z}_0||^2) \right].
\end{aligned}
\end{equation}

\subsection{Sampling stage}
DPM-solver++~\citep{lu2022dpm, lu2022dpm++} is proposed totally based on the continuous diffusion models. According to it, we have the exact solution of diffusion ODEs, given an initial value $\mathbf{z}_s$ at time $s > 0$:
\begin{equation}
    \mathbf{z}_t = \frac{\sigma_t}{\sigma_s} \mathbf{z}_s + \sigma_t \int_{\lambda_s}^{\lambda_t} e^{\lambda} f_\theta \,d\lambda,
\end{equation}
where the $\lambda$ is a strictly decreasing function of $t$, $\sigma_t$ is proportional to $\beta_t$, specifically, $\sigma_t=\sqrt{1-\bar{\alpha}_t}$.

We need to approximate $\int e^{\lambda} f_{\theta}\,d\lambda$, which can be analytically computed by repeatedly applying $n$ times of integration-by-parts. According to the second order multistep DPM-Solver++ algorithm, the reconstruct of $\mathbf{z}_0$ relies on the $f_\theta$, after posing discrete denoise in our methods, the algorithm still applies since our $f_\theta(\mathbf{\hat{z}}_\lambda, \lambda)$ is exactly aligned with training objectives. 

\section{Related Work}
\label{sec:related-work}
Continuous diffusion models are first applied in image generation~\citep{song2020denoising, ho2020denoising} and then applied in text generation~\citep{li2022diffusion, gong2022diffuseq}. Meanwhile, discrete diffusion models~\cite{austin2021structured, hoogeboom2021argmax, Zheng2023ARD} is designed for text generation. \citep{he2022diffusionbert} and \citep{Zhou2023DiffusionNATSD} directly leverage the [MASK] token used in pre-trained language models. By contrast, our method learns the soft absorbing state from scratch along with the whole diffusion process and bridges discrete diffusion with continuous space. The idea of absorbing state or discretely corrupt data can be seen in many NLP work like BERT~\citep{he2022diffusionbert} or BART~\citep{lewis2020bart}, or even in diffusion image generation~\citep{hu2022unified}.

\section{Implementation Details}
\subsection{General setting}
We use 2 A100 80G GPUs for training using FP16 with a batch size of 425 and a single GPU on sampling with a batch size of 100. The generation speed is measured under the setting of 100 batch size on one NVIDIA A100 80G GPU for all models, averaged by 3 runs. 

To evaluate the quality, we use the standard metric BLEU~\citep{papineni-etal-2002-bleu} and ROUGE~\citep{lin2004rouge} score. 
Since string-similarity-based metrics can be unsatisfactory for open-ended generation, we also report BERTScore~\citep{zhang2019bertscore} that assesses the semantic similarity between generated sentences and references.
For sentence-level diversity evaluation, we consider sentence-level self-BLEU~\citep{zhu2018texygen} to measure the n-gram overlap between the set of outputs w.r.t one source sentence. The self-BLEU score is computed using 2 samples for each test case generated with different seeds. All ablation study is conducted using the original version of our models if not specified, to validate the effectiveness of introducing the soft absorbing state.

\subsection{Baselines setting}
All baselines we used are open-sourced. DiffuSeq\footnote{\url{https://github.com/Shark-NLP/DiffuSeq}}~\citep{gong2022diffuseq} and BG-DiffuSeq\footnote{\url{https://github.com/CODINNLG/Bridge\_Gap\_Diffusion}}~\citep{Tang2023CanDM} are implemented based on HuggingFace \texttt{Transformers}\footnote{\url{https://github.com/huggingface/transformers}}.
RDM\footnote{\url{https://github.com/hkunlp/reparam-discrete-diffusion}}~\citep{Zheng2023ARD} are implemented using \texttt{Fairseq}, where the temperature sampling is adopted when sampling and beam size is set to 1. For generation speed, we report the results on a single NVIDIA A100 80G GPU with batch size 100.
\label{sec:appendix details}

\end{document}